\pdfoutput=1

\documentclass[11pt]{article}

\usepackage[]{acl}
 
\usepackage{times}
\usepackage{latexsym}

\usepackage{graphicx}
\usepackage{comment}
\usepackage{multirow}
\usepackage{diagbox}
\usepackage{slashbox}
\usepackage{float}
\usepackage{afterpage}
\usepackage{amssymb}
\usepackage{commath}
\usepackage{amsmath}

\usepackage{array}
\newsavebox\dummy
\newcolumntype{H}{>{\begin{lrbox}{\dummy}}c<{\end{lrbox}}@{}}

\usepackage[T1]{fontenc}

\usepackage[utf8]{inputenc}

\usepackage{microtype}

\title{QaNER: Prompting Question Answering Models\\for Few-shot Named Entity Recognition}

\author{Andy T. Liu$^{\ddagger \mathparagraph}$\thanks{$\;\;$Work done while interning at Amazon AI.}, Wei Xiao$^{\ddagger}$, Henghui Zhu$^{\ddagger}$ \\
{\bf Dejiao Zhang$^{\ddagger}$, Shang-Wen~Li$^{\mathsection}$\thanks{$\;\;$Work done while working at Amazon AI.}, Andrew Arnold$^{\ddagger}$} \\
$^{\ddagger}$Amazon AI, USA, $^{\mathsection}$Facebook AI \and $^{\mathparagraph}$National Taiwan University, Taiwan \\
\texttt{liuandyt@gmail.com, shangwel@fb.com} \\ \texttt{\{weixiaow, henghui, dejiaoz, anarnld\}@amazon.com}}

\begin{document}

\maketitle
\begin{abstract}
Recently, prompt-based learning for pre-trained language models has succeeded in few-shot Named Entity Recognition (NER) by exploiting prompts as task guidance to increase label efficiency. However, previous prompt-based methods for few-shot NER have limitations such as a higher computational complexity, poor zero-shot ability, requiring manual prompt engineering, or lack of prompt robustness. In this work, we address these shortcomings by proposing a new prompt-based learning NER method with Question Answering (QA), called QaNER. Our approach includes 1) a refined strategy for converting NER problems into the QA formulation; 2) NER prompt generation for QA models; 3) prompt-based tuning with QA models on a few annotated NER examples; 4) zero-shot NER by prompting the QA model. Comparing the proposed approach with previous methods, QaNER is faster at inference, insensitive to the prompt quality, and robust to hyper-parameters, as well as demonstrating significantly better low-resource performance and zero-shot capability.
\end{abstract}

\section{Introduction}
Named Entity Recognition (NER) aims to tag entities in text with their corresponding type.
In previous works~\citep{lstm, conll03, bert, supervised-ner-crf}, the NER problem is often formulated as a sequence labeling problem (also known as sequence tagging or token classification). 
With supervised learning, each entity in the text sequence can be assigned to a pre-defined entity label.
Training a supervised NER system requires many labeled training data.
However, labeling a large corpus of tokens requires deep domain knowledge, 
which makes creating such a corpus expensive and time-consuming.
Also, building NER systems at scale with rich annotations for different real-world application scenarios is labor-intensive work. 
There are potentially hundreds of new domains for enterprise use-cases, not to mention different languages.
These reasons motivate a practical and challenging research problem: few-shot NER~\citep{few-shot-ner}.

\begin{figure}[t]
\centering
\includegraphics[width=\linewidth]{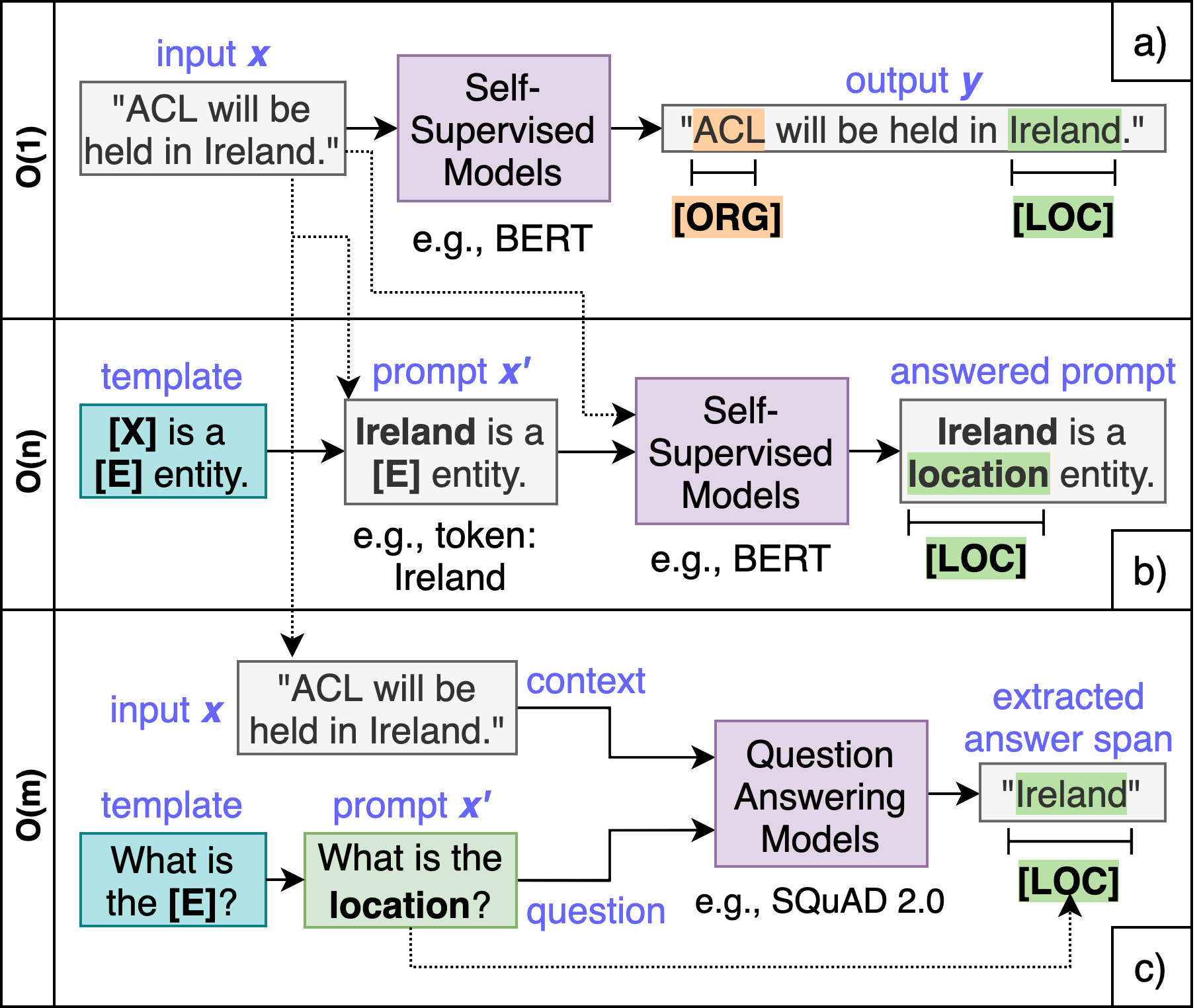}
\caption{Different learning styles for NER: a) Sequence labeling; b) Prompt-based learning with LM; and c) Proposed: Prompt-based learning with QA Models. 
For Big-O notations, $n$ is the length of the input $x$, and $m$ is the number of entity types. Note that $m$ is usually significantly lower than $n$.}
\label{fig:proposed}
\end{figure}

Meanwhile, learning with prompt-based tuning is a new paradigm in the Natural Language Processing (NLP) community with growing popularity~\citep{prompt-survey}.
Prompt-based methods reformulate the input to Language Models (LM) to bridge the gap between pre-training and downstream tasks.
Previous work~\citep{unified-mrc} has formulated the task of NER as a machine reading comprehension (MRC) task.
However, they did not study the design of prompts (questions for the MRC models to answer), nor did they experiment with few-shot or zero-shot settings.
In our work, we show that QA models can leverage knowledge learned from question-answering data to improve low-resource NER performance.
We achieve better results in few-shot and zero-shot scenarios because prompt-based learning can increase label efficiency (i.e., adapting to a new domain with fewer examples).

There are four limitations regarding previous prompt-based NER methods.  
First, the inference time is in proportion to the length of the sequence~\citep{light-ner}, or requires one prompt per token~\citep{template-ner}. Thus, existing methods have \textbf{high computational complexity}.
Second, previous prompt-based methods~\citep{template-ner} \textbf{require manual prompt engineering} to design a template, which is a labor-intensive process. 
Third, prompt-based NER~\citep{template-ner} tends to \textbf{lack prompt robustness}. The method is sensitive to different prompt designs and depends on tuning with a development set of significant size, which may be unavailable in low-resourced cases. 
Lastly, previous methods~\citep{template-ner, light-ner} use a high-resource NER dataset to transfer knowledge for few-shot learning. 
The performance is impaired as datasets employ different sets of labels, and hence the methods have \textbf{less flexible transferability}.

To address these challenges, we propose to prompt off-the-shelf Question Answering (QA) models for NER.
As illustrated in Figure 1 (c), our approach is based on the idea that if we can formulate an NER problem as a QA task, a QA model should answer the question and solve the original problem.
We evaluate our method on NER datasets of CoNLL03~\citep{conll03}, MIT Movie Review~\citep{mit-datasets}, and MIT Restaurant Review~\citep{mit-datasets}.
Our method empirically shows significantly better performance over previous approaches, especially when labeling resources are low.
The contributions of our paper are as follows:

1) We are the first work to introduce a prompt-guided Question Answering framework for NER (QaNER). 
QaNER achieves state-of-the-art performance in low-resource settings and competitive results when trained with full datasets. 
The improvement largely comes from the proper knowledge transfer from QA models.

2) We improve the method's robustness against prompt design and development set size with QaNER.
QaNER continues to perform well with various prompts (prompt generation with pre-trained LM) and when there is no development set for parameter tuning to hypothesize (the "shoot in the dark" setting).
Such robustness can be attributed to the significant naturalness of the formulation of the NER problem into QA.

3) QaNER improves the computational complexity and achieves better knowledge transferability since asking QA models questions is very natural, as prompting QA models introduces less mismatch when compared to prompting other LMs.
The QA problem formulation allows us to identify all entity spans for each type with one inference.

\begin{table*}[ht]
\centering
\begin{tabular}{|l||l|l|l|}
\hline
Method + Model & Objective & Complexity of Model & Complexity of Method\\ \hline \hline
Sequence Labeling BERT & classification & $O(\;n^{2}d+nd^{2})$ & $O(1)$ \\ \hline
TemplateNER BART & generation & $O((n^{2}d+nd^{2}){\times}(t+1))$ & $O((m+1){\times}n\hat{n})$ \\ \hline
LightNER BART & generation & $O((n^{2}d+nd^{2}){\times}(t+1))$ & $O(1)$ \\ \hline
QaNER BERT (Ours) & extraction & $O(\;n^{2}d+nd^{2})$ & $O(m)$ \\ \hline
\end{tabular}
\caption{We show the formulations of different NER models, where $n$ denotes the length of input; $d$ the dimension of models; $t$ the amount of autoregressive steps; $\hat{n}$ implies $\hat{n}$-grams; and $m$ the number of entity types. TemplateNER prompts with an additional none entity type hence the $O(m+1)$.}
\label{tab:complexity}
\end{table*}

\section{Background}

\subsection{Few shot Named Entity Recognition}
\label{ssec:few-shot-ner}
In previous works~\citep{few-shot-ner, template-ner, example-bert, light-ner}, a commonly used approach is to assume a high-resource NER dataset, where there is a large number of training instances. 
The model is first trained on the source dataset with high resource and then transferred to the target low-resource NER dataset.
The two datasets may have different domains, i.e., the entity types are different between the high-resource source dataset and low-resource target dataset.

\subsection{True Low-resource Learning}
A constantly neglected problem in low-resource NER studies is to assume and exploit a large development set (dev set).
The work of \citealt{true-few-shot} suggests that prior works significantly overestimated the true few-shot ability of their models, as many held-out examples are used for various aspects of learning.
There are several different approaches to account for this problem.

\textbf{High resource dev set.} 
The dev set from the high resource dataset 
is used to tune hyper-parameters and choose templates (“prompts”). 
This approach assumes a large dev set for the task but on a different domain~\citep{template-ner, light-ner}.
\textbf{Small dev set.} In \citealt{lmbff}, a few-shot dev set with the same size as the training set is randomly sampled, keeping the setting "few-shot" while still able to tune hyper-parameters and choose templates.
\textbf{No dev set.} In the work of \citealt{size-not-matter} and \citealt{true-few-shot}, they choose not to use any development data and adopt fixed hyper-parameters, where models are evaluated when held-out examples are unavailable. 
This setting is also known as ``shooting in the dark'' or ``true few-shot learning''~\citep{true-few-shot}.


\subsection{Related Work}
\label{ssec:related-work}
\subsubsection{Sequence Labeling}
Traditionally, the NER task is regarded as a sequence labeling problem.
Pre-trained models like BERT~\citep{bert} are used as the encoder for input $x_{1:n}$ which generates sequence representations $h_{1:n}$, where $n$ is the number of tokens in the input $x$.
A classifier is trained on top of the encoder model to map hidden representations to token labels as $P(y_{1:n}|x_{1:n})$.
In Figure~\ref{fig:proposed} a), we illustrate sequence labeling NER.

\subsubsection{Prompt-based Learning for NER}
Recently, prompt-based learning has emerged to bridge the gap of mismatch in pre-training and fine-tuning.
Applying prompt-based learning to NER, the original input $x$ is first modified into a \textit{template}, which results in a new textual string with some unfilled slots called \textit{prompt} $x^{\prime}$.
Then a LM is used to fill the unfilled slots in $x^{\prime}$ to obtain the final string $\hat{x}$, and finally, from $\hat{x}$ we can derive the output answer $y$.
In Figure~\ref{fig:proposed} b), we illustrate the simplified heuristic of previous methods, which has method complexity of $O(n)$, where $n$ denotes the length of the input.

\textbf{TemplateNER.} 
In \citealt{template-ner}, a template-based method is introduced where BART~\citep{bart} is used as the backbone.
In Figure~\ref{fig:proposed} b), the simplified version, we assume the \texttt{[E]} slot can only be filled with tokens.
However, in the actual case, \texttt{[E]} can be the enumeration of all possible spans of $\hat{n}$-grams in the sentence.
Given an input sentence of length $n$, this results in a complexity of $O(n{\times}\hat{n})$.
The enumerated spans are filled in the handcrafted templates, where there are $m$ different prompts corresponding to different entity types, plus an additional prompt used for the none entity type, and $m$ is the number of entity types in a dataset.
Hence the TemplateNER method has a complexity of $O((m+1){\times}n\hat{n})$.

\textbf{LightNER.}
In \citealt{light-ner} the NER task is formulated as a generation problem.
LightNER also adopts BART~\citep{bart} as the backbone model, which generates the index of the entity span in the input as well as the entity type labels.
The complexity of the LightNER method is $O(1)$, as the output at each generative step is the prediction of entity span along with entity type.

\subsection{Complexity}
To understand the overall complexity of each method, here we introduce the complexity of backbone Transformer~\citep{transformer} models.
The complexity of the transformer encoder is $O(n^{2}d+nd^{2})$~\citep{transformer, accelerate-transformer}, where $d$ is the dimension of the model.
The transformer decoder performs multi-head
attention over the output of the encoder stack for $t$ autoregressive steps, and thus has the complexity of $O((n^{2}d+nd^{2}){\times}t)$.
For models like BART~\citep{bart}, that use the encoder and the autoregressive decoder, the overall complexity of model is $O((n^{2}d+nd^{2}){\times}(t+1))$.
We summarize the complexity of model together with complexity of method (as discussed in Section~\ref{ssec:related-work}) in Table~\ref{tab:complexity}.
Thus the overall complexity is the multiplication of "complexity of model" and "complexity of method".
In general $m$ is much smaller than $t$, thus making QaNER faster than LightNER in most cases.

\section{Methodology}

\subsection{Extractive QA}
Among a variety of different QA formats~\citep{unified-qa}, we consider the formulation of extractive QA (span-based QA).
The choice of extractive QA comes naturally as it perfectly fits the NER objective to recognize entity spans.
In extractive QA, given a question $Q$ and a context of text $C$ that might contain the answer, the model needs to extract the corresponding answer $A$ as a sub-string of $C$.
Thus each instance in a QA dataset is a tuple of $(C, Q, A)$.
In some datasets like SQuAD 2.0~\citep{squad2}, "unanswerable" might be the correct response for unanswerable questions, and the model extracts the special token "\texttt{[CLS]}".

In this work, we use BERT~\citep{bert} as the backbone model for QA, to encode the representation of text sequence $x_{1:n}$ to $h_{1:n}$, where $x_{1:n}$ is the concatenation of question $Q$ and context $C$.
On top of BERT, a QA head maps hidden representations to the predicted span.
The QA head is essentially a start and end token classifier, where one prediction marks the start of the answer and the other marks the end.
Using extractive QA models, we treat NER as a span-based extraction problem. 
The prediction per entity type is based on each token's start/end scores, which is essentially the same as the NER objective, to locate the start and end index that marks the entity span from the input, along with the entity type.
In this work, we adopt off-the-shelf BERT Large models fine-tuned on SQuAD or SQuAD 2.0~\citep{squad, squad2} for our prompt-based scheme.

\subsection{Prompting the QA Model}
\label{ssec:prompt-qa}
Prompting the QA model is intuitive.
As a result, it takes little effort in the template design.
Using naive questions (\textit{templates}) like "What is the \texttt{[E]}?" will simply work, where $[\texttt{E}] \in \{``\texttt{person}", ``\texttt{location}", \cdot\}$ is the set of entity types in the dataset.
Given a fixed \textit{template}, we fill the \texttt{[E]} slot with an entity type to create a new textual string called \textit{prompt} $x^{\prime}$.
Hence we generate $m$ prompts for $m$ entity types.
We apply some normalization for non-alphabetic characters in the entity type. For example, we convert \texttt{restaurant\_name} to \texttt{restaurant name} by replacing underscore ($\_$) with a space.

\subsection{Prompt Generation with Pre-trained LM}
We explore adding several variations in generating the prompt by using an MLM (Masked Language Models)~\citep{bert} to fill the masked token in the template "\texttt{[MASK]} is the \texttt{[E]}?" after lexicalizing the \texttt{[E]} slot.
For example, we take entity type "location" and insert it into template to obtain "\texttt{[MASK]} is the location?". The MLM is then used to fill the mask and generate the prompt: "Where is the location?".
The only difference between handcrafted and generated prompts is that humans decide the \texttt{[MASK]} token in the former prompt, and the other is decided by the MLM.
We use another BERT model to generate the prompts in advance.
We observe some differences in human-filled handcrafted prompts and BERT-filled prompts. 
For the same example, "\texttt{[MASK]} is the location?", a handcrafted prompt filled by a human would be "What is the location?" rather than "Where is the location?".
The former is more like a QA question (as perceived by a human), and the latter comes more naturally to the MLM.
Our experiments also study more template designs with other question formulations, for example, "Is there a \texttt{[E]}?".
Empirically, we find that prompts with the \textit{"five Ws"} (Who, What, When Where, and Why) question words worked the best in general, as they match the formulation of QA questions.
Considering that for each entity type, one corresponding prompt sentence needs to be generated, thus automatic prompt generation can be helpful for datasets that have a large number of entity types, for example, the Ultra-Fine dataset~\citep{ultra-fine}.

\begin{figure}[t]
\centering
\includegraphics[width=\linewidth]{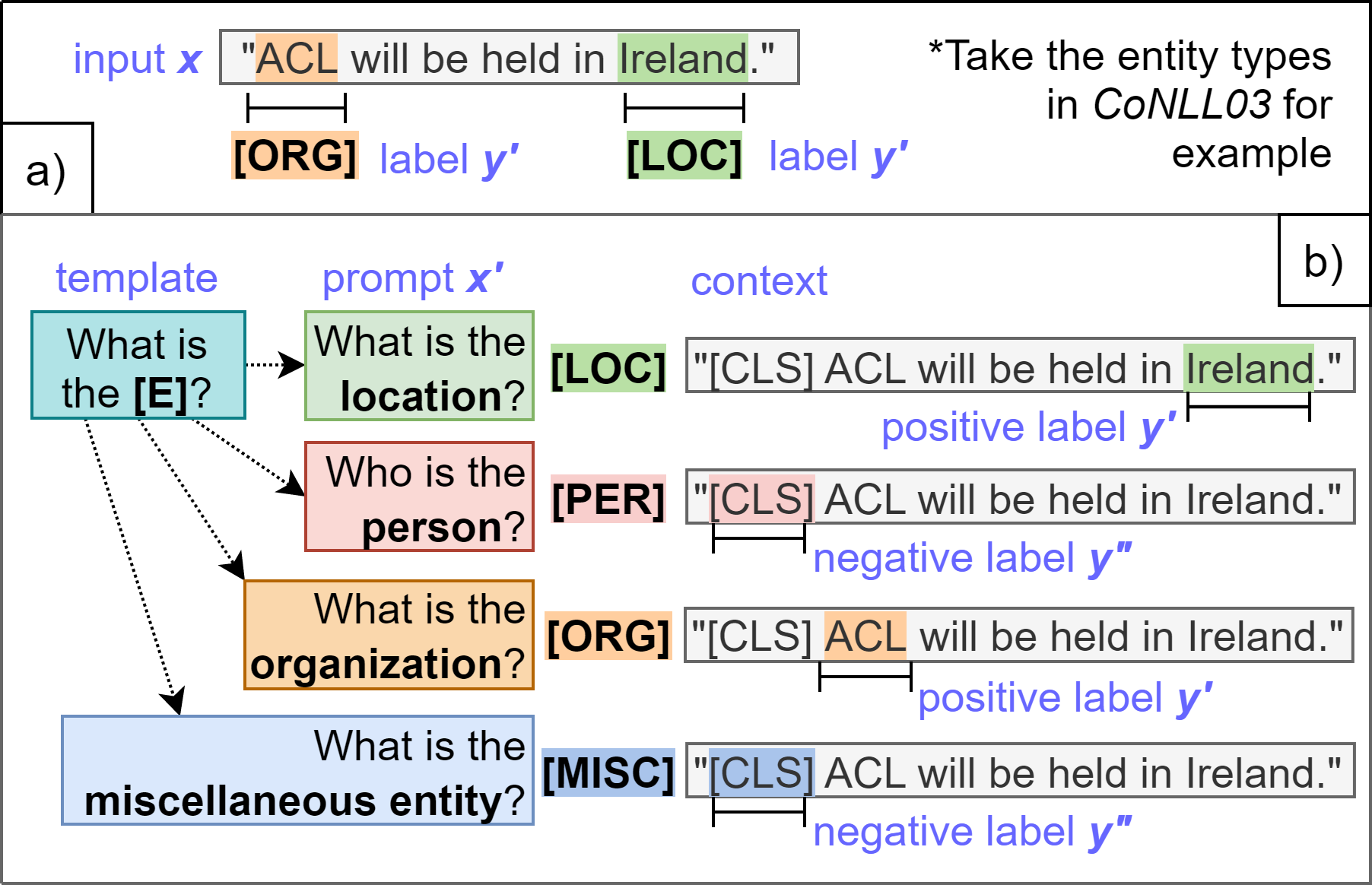}
\caption{Illustration of an NER example in a) transform into the QA formulation in b).}
\label{fig:convert}
\end{figure}

\subsection{Converting NER to QA}
\subsubsection{Positive and Negative Examples}
For each NER instance, we generate $m$ prompts as described in the previous Section~\ref{ssec:prompt-qa}, and match these prompts with their corresponding answer span as shown in Figure~\ref{fig:convert}.
In this work, we refer to training instances with answerable questions as \textit{positive examples}.
For the unanswerable questions, we mark the special token "\texttt{[CLS]}" as the answer, following the SQuAD 2.0~\citep{squad2} data format.
We refer to these impossible examples as \textit{negative examples}.
The \textit{negative examples} serve to help the model identify \textit{positive examples} correctly, similar to the idea in contrastive learning~\citep{cpc}.
Our experiments in Section~\ref{ssec:ablation} show the importance of training with negative examples, which does not largely affect question answering performance but is crucial in our prompt-based tuning scheme.

\subsubsection{Repeating Examples}
In an NER instance, for any entity type $e$, $e$ may show up more than once, and at varying locations in the input $x_{1:n}$.
For example, there may easily be more than one $PER$ type in a sentence.
Theoretically, there can be at most $n$ separate entity labels while the entity amount $m$ is fixed (where $m$ is also the number of prompts for our method).
To account for situations like this, we allow \textit{repeating examples} during the NER to QA conversion process.
In other words, for the same context and question pair $(C, Q)$, there may be different answers $A_i$ such that the QA instance is $(C, Q, A_i)$, where $i \in {1, 2, ..., I}$ and $I$ is the number of time a particular entity type $e$ repeats in an NER example $x_{1:n}$.

To efficiently retrieve all the tokens of the same entity type (but at different input locations) in one pass,
we first fine-tune the QA model to recognize entity types at different locations.
Then during the QA decoding process, we retrieve the n-best candidate results (by computing over the start/end scores of every token). 
This allows us to identify all the tokens corresponding to the target entity type in a single prompt inference.
If two extracted spans overlap, we choose the span with the higher score to avoid possible prediction contradictions.

\begin{figure*}[ht]
\centering
\includegraphics[width=\linewidth]{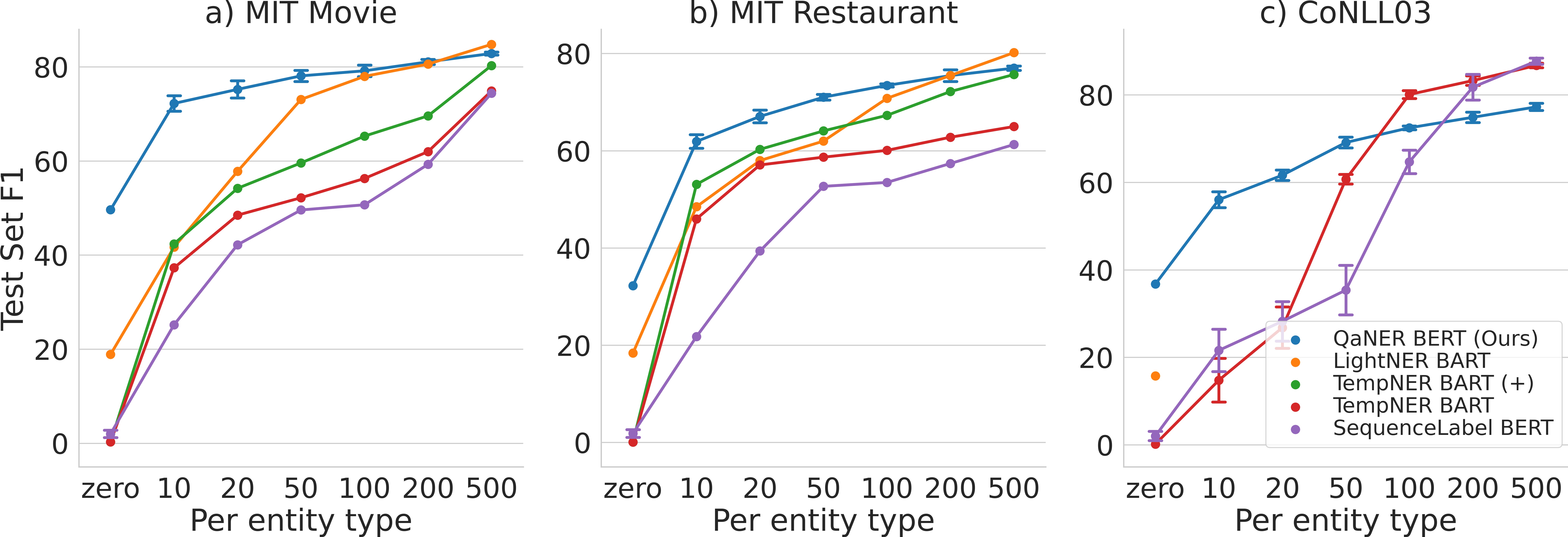}
\caption{Few-shot NER performance on different test sets. (+) indicates training on extra NER data - CoNLL03. N per entity type indicates N instances for each entity type. For LightNER BART and TemplateNER BART, in all their zero-shot settings, we used extra NER data from the other two datasets for cross-domain zero-shot. (i.e., zero-shot on MIT Movie with all data from MIT Restaurant + CoNLL03, and so on.)}
\label{fig:exp-main}
\end{figure*}

\begin{figure*}[t]
\centering
\includegraphics[width=\linewidth]{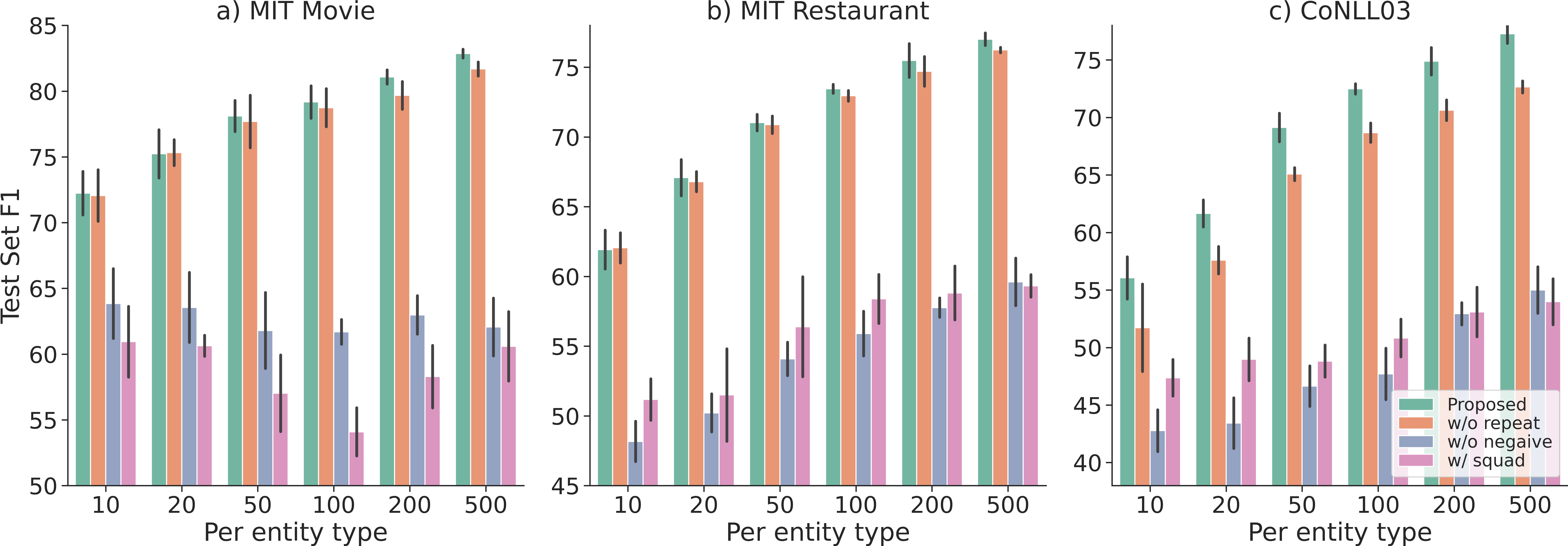}
\caption{Ablation study - we show the importance of using \textit{repeating examples} and \textit{negative examples}. }
\label{fig:exp-ablation}
\end{figure*}

\section{Experimental Setup}

\subsection{Dataset Setup}
In our experiments, we consider three datasets for NER, namely MIT Movie, MIT Restaurant~\citep{mit-datasets}, and the CoNLL03~\citep{conll03} dataset.
\textbf{The MIT Movie Dataset} and \textbf{the MIT Restaurant Dataset}
are semantically tagged training and test corpus in BIO format, which contain user queries about movie or restaurant information~\citep{mit-datasets}.
Since MIT Movie and Restaurant datasets contain no dev set, we randomly sample and isolate 10\% of the training set as dev set for experimental purposes.
\textbf{The CoNLL03 Dataset}
is an NER dataset released as a part of CoNLL-2003 shared task~\citep{conll03}. 
The dataset contains training, development, and test sets.
In this paper, we use the English version which was taken from the Reuters Corpus.

To evaluate few-shot performance on NER datasets, we adopt the commonly used \textit{N per entity type} setting, where \textit{N} indicates \textit{N} instances for each entity type.
We randomly sample \textit{N} instances per entity category from the dataset, and set \textit{N} to 10, 20, 50, 100, 200, and 500 following previous works~\citep{template-ner, light-ner, example-bert}.
We strictly constrain at most N instances per entity type. 
As a result, we may not have N for all types. 
Some types may have fewer than N instances.

For QA models, we consider two datasets.
\textbf{The SQuAD Dataset}
where the answer to every question is a segment of text (or span) from the context~\citep{squad}.
\textbf{The SQuAD 2.0 Dataset} combines the 100,000 questions in SQuAD with an additional over 50,000 unanswerable questions~\citep{squad2}.
The unanswerable questions allow us to use negative examples in our prompt-based QA fine-tuning scheme, as shown in Figure~\ref{fig:convert}.
In our experiments, the proposed method adopts QA models fine-tuned on SQuAD 2.0 if not mentioned explicitly.

\subsection{Development Set Design}
Unlike previous works~\citep{template-ner, light-ner}, we do not rely on high resource dev sets, which essentially are large NER dev sets from a different domain.
Instead, we design two practical dev sets to evaluate the true low-resource performance fairly.
1) we use a randomly sampled \textit{small dev set}, with the \textbf{same size} as the smallest training set (the \textit{10 per entity type} setting), thus keeping this dev set consistently small.
2) we use a \textit{10 per type dev set}, where we randomly sample 10 instances per entity category from the entire dev set.
This setting ensures that there is a sufficient amount of dev data for each entity type and allows us to study the effect of a not balanced dev set (by comparing \textit{10 per type dev set} to \textit{small dev set}).

In addition, we also adopt 3) the \textit{no dev set} setting, following \citealt{size-not-matter, true-few-shot}, where we do not use any development data and adopt fixed hyper-parameters.
For this setting, 
we adopt out-of-the-box hyper-parameters as suggested in the BERT paper~\citep{bert}.
Specifically, when fine-tuning QA models on NER with \textit{no dev set}, we use a learning rate of $2e^{-5}$, batch size of $16$, and tune for $4$ epochs.
The QA decoding process uses an n-best size of $20$, a maximum answer length up to $30$, and a threshold that requires the sum of predicted start and end token probability to exceed $100\%$.
In our experiments, the proposed method adopts the \textit{no dev set} setting by default if not mentioned explicitly.

\subsection{Evaluation Protocol}
It is well known that fine-tuning on a small number of examples can suffer from instability~\citep{5-random-example, 5-random-example-2}, and results may change dramatically for a different split of data. 
To account for this, in all of our few-shot experiments, we measure average performance across five different randomly sampled splits of the training set.
Sampling multiple random splits gives a more robust measure of the performance and variance.

\begin{figure*}[ht]
\centering
\includegraphics[width=\linewidth]{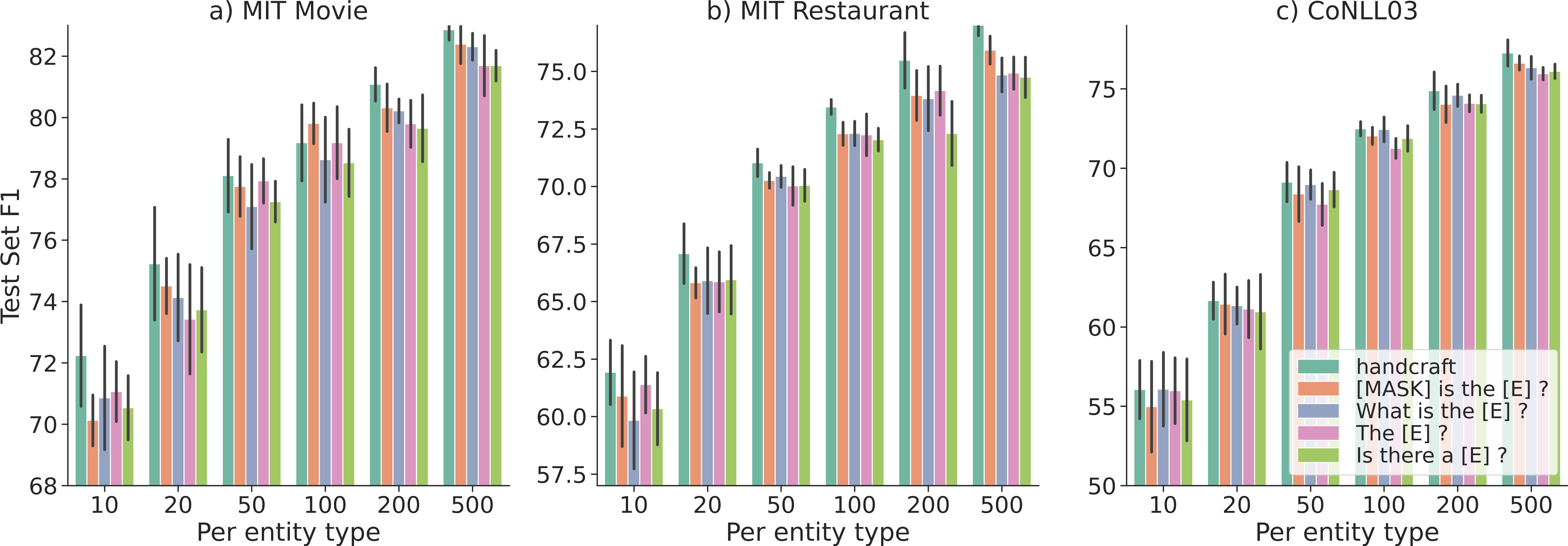}
\caption{The effect of different templates, using "\texttt{[MASK]} is the \texttt{[E]}?" yields close performance to \textit{handcraft}. }
\label{fig:exp-template}
\end{figure*}

\begin{table*}[ht]
\centering
\begin{tabular}{|l||c|c|c||c|c|c||c|}
\hline
\multirow{2}{*}{\backslashbox{Template}{Data}} & \multicolumn{3}{c||}{zero-shot} & \multicolumn{3}{c||}{full resource} & \multirow{2}{*}{Rank} \\ \cline{2-7}
& a) Mov & b) Rest & c) CoN & a) Mov & b) Rest & c) CoN & \\ \hline \hline
handcraft & 49.65$_1$ & 32.25$_1$ & 36.77$_2$ & 85.53$_1$ & 79.57$_1$ & 80.92$_1$ & 1.17$_1$ \\ \hline
{[}MASK{]} is the \textless{}e\textgreater ? & 44.98$_3$ & 31.29$_2$ & 37.00$_1$ & 85.10$_2$ & 77.79$_2$ & 79.20$_3$ & 1.86$_2$ \\ \hline
What is the \textless{}e\textgreater ? & 49.34$_2$ & 29.46$_3$ & 35.89$_3$ & 84.82$_4$ & 77.32$_4$ & 78.39$_5$ & 3.50$_3$ \\ \hline
The \textless{}e\textgreater ? & 34.35$_4$ & 14.93$_4$ & 22.81$_4$ & 84.95$_3$ & 77.53$_3$ & 79.19$_4$ & 3.67$_4$ \\ \hline
Is there a \textless{}e\textgreater ? & 31.14$_5$ & 14.55$_5$ & 20.60$_5$ & 84.01$_5$ & 76.71$_5$ & 79.71$_2$ & 4.50$_5$ \\ \hline
\end{tabular}
\caption{The effect of different templates in full resource and zero-shot settings. The subscripts indicate the ranking in each column, and the rightmost column shows the average ranking of each row.}
\label{tab:exp-template}
\end{table*}

\begin{figure*}[ht]
\centering
\includegraphics[width=\linewidth]{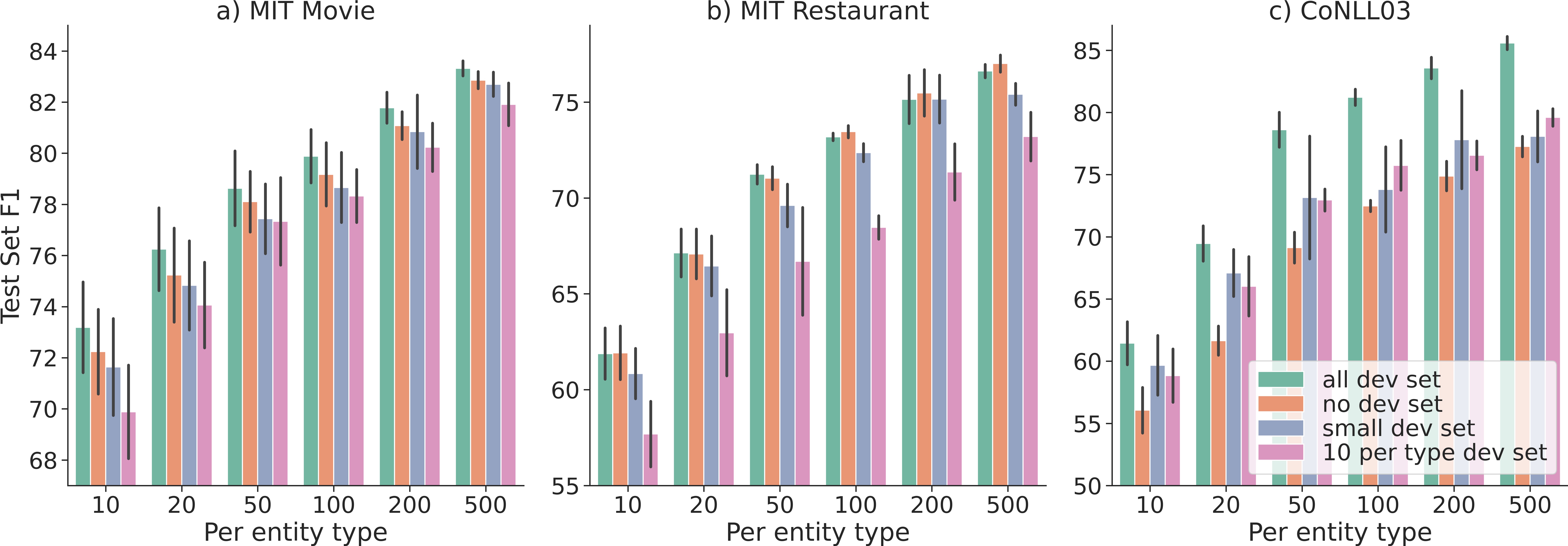}
\caption{The effect of different development set sizes in the few-shot setting. }
\label{fig:exp-dev}
\end{figure*}

\begin{table*}[ht]
\centering
\begin{tabular}{|l||c|c|c||c|c|c||c|}
\hline
\multirow{2}{*}{\backslashbox{Template}{Data}} & \multicolumn{3}{c||}{zero-shot} & \multicolumn{3}{c||}{full resource} & \multirow{2}{*}{Rank} \\ \cline{2-7}
& a) Mov & b) Rest & c) CoN & a) Mov & b) Rest & c) CoN & \\ \hline \hline
all dev set & 56.20$_1$ & 31.27$_1$ & 37.80$_2$ & 85.36$_1$ & 78.86$_1$ & 89.04$_1$ & 1.17$_1$ \\ \hline
no dev set & 48.72$_2$ & 29.78$_3$ & 36.77$_3$ & 84.55$_2$ & 77.40$_3$ & 80.92$_3$ & 2.67$_2$ \\ \hline
small dev set & 41.75$_3$ & 30.87$_2$ & 37.89$_1$ & 85.14$_3$ & 76.95$_4$ & 78.26$_4$ & 2.83$_3$ \\ \hline
10 per type & 36.15$_4$ & 29.64$_4$ & 33.72$_4$ & 84.25$_4$ & 78.44$_2$ & 84.13$_2$ & 3.33$_4$ \\ \hline
\end{tabular}
\caption{The effect of different development set sizes in full resource and zero-shot settings. The subscripts indicate the ranking in each column, and the rightmost column shows the average ranking of each row.}
\label{tab:exp-dev}
\end{table*}

\section{Results}

\subsection{Comparison of Recent Approaches}
We compare the performance of different methods in Figure~\ref{fig:exp-main}, where we show the performance variation when training with different amounts of training data.
Here we use handcrafted templates on QaNER, to match the setting of other work. 
The performance of generated templates will be shown in later sections.
In terms of zero-shot performance, the proposed method significantly outperforms previous works, despite their using extra NER data from high resource datasets.
Both TemplateNER~\citep{template-ner} and LightNER~\citep{light-ner} claim that their methods possess zero-shot ability.
We show that the proposed method has improved zero-shot performance.
In terms of few-shot performance, the proposed method has a notable advantage when the number of used labeled examples is low (10, 20, 50 per entity type).
As training data increases, the gap between the proposed method and other methods closes.
The reason is that the advantage of exploiting QA knowledge has been dominated by NER labeled data as more labeled examples are used.
The gap closes faster for easier NER datasets. 
For example, CoNLL03 only has four entity types, making it easier for other methods to master.
On the other hand, MIT Movie and MIT Restaurant have more entity types and are more challenging for learning. 
Hence QaNER continues leading in performance as more data are used.
In addition, the proposed method adopts \textit{no dev set} in the above comparison, whereas TemplateNER~\citep{template-ner} and LightNER~\citep{light-ner} tune their parameters on the high resource dev set.
The performance of the proposed method can be improved when tuned with development data, as we show in a later section.

\subsection{Ablation Study}
\label{ssec:ablation}
In this section, we study the components of the proposed method.
In Figure~\ref{fig:exp-ablation}, we study the differences among the components of the proposed method, without \textit{repeating examples}, without \textit{negative examples}, and using SQuAD instead of SQuAD 2.0 for the training of QA models.
In Figure~\ref{fig:exp-ablation}, all the ablated methods are with handcrafted templates to reduce variation. 
First, we observe that including \textit{repeating examples} slightly improves the model's performance on the MIT Movie and Restaurant datasets, while having a more significant on CoNLL03.
The reason is that there are more repeating instances in the CoNLL03 dataset.
Secondly from comparison we see that adopting \textit{negative examples} makes a large difference, as \textit{negative examples} help the model to recognize \textit{positive examples} correctly.
Thirdly, comparing to \textit{w/ SQuAD}, we note that SQuAD 2.0 is not beneficial without fine-tuning with negative NER examples.

\subsection{Different Prompt Designs}
In Figure~\ref{fig:exp-template}, we study the effect of adopting different template designs.
Here we tried several ways to generate the prompts and the results are quite consistent, which shows the robustness of methods (even at few-shot cases).
In addition, zero-shot and full resource performance are shown in Table~\ref{tab:exp-template}.
The zero-shot performance shows the "naturalness" of the prompt regarding the QA model, and the full resource performance is helpful to understand each prompt's capability.
We see handcraft prompts outperform the others, followed by generated prompts ("[MASK] is the \texttt{[E]}?").
In "What is the \texttt{[E]}?", we fix all the \textit{``Ws"} to "What" and get slightly lower performance than generated prompts.
In "The \texttt{[E]}?", we drop the \textit{``Ws"} and get degraded performance.
We learn that choosing the appropriate \textit{``five Ws"} can influence the QA model's performance by a small margin.
In "Is there a \texttt{[E]}?", we use a different type of question, without \textit{``Ws"} and get the lowest result on average.
We conclude that prompts with the \textit{``five Ws"} work the best in general, especially handcrafted and generated prompts, as they match the modeling of QA.
Furthermore, this study shows that QaNER is robust against the design of prompts.

\subsection{Development Set Study}
In Figure~\ref{fig:exp-dev}, we study the effect of adopting different development set settings.
Also, we show their zero-shot and full resource performance in Table~\ref{tab:exp-dev}.
We report the \textit{all dev set} setting to serve as a top-line, as it is not realistic to assume the existence of such large dev sets in low-resource scenarios.
We observe that, on average, using \textit{no dev set} outperforms other dev set settings, in particular on the MIT Movie and Restaurant datasets.
The reason is that the CoNLL03 dataset has a standard development set that matches the distribution of the test set.
On the other hand, the MIT datasets use dev sets randomly sampled from the training data. 
Using those dev sets may result in overfitting.
Note that previous works tune with the CoNLL03 standard and large dev set, while we use \textit{no dev set} when comparing with them in Figure~\ref{fig:exp-main}.
In other words, we report the lower-bound performance of the proposed method in Figure~\ref{fig:exp-main}, as QaNER was not tuned on the CoNLL03 dataset.
Using \textit{small dev set} and \textit{10 per type dev} set achieves worse performance when compared to \textit{no dev set} on the MIT datasets.
However, this is not the case for CoNLL03. The reason is the same as previously discussed.
In general, assuming we do not know the distribution of dev and test sets, using \textit{no dev set} yields the most robust performance.
To conclude this section, QaNER works well even when "shooting-in-the-dark" and is robust against different dev set settings.

\section{Conclusion}
In this work, we propose QaNER, where we prompt the QA model for the NER task. 
We demonstrate how to convert NER examples into QA examples and how to perform prompt generation for QA models. 
QaNER is not only faster and more robust in low-resource conditions but also promising in few-shot and especially zero-shot performance, opening a new door for self-learning methods.
In future work, we aim to explore prompting the QA model for other NLP tasks, including sentence classification and extractive summarization.


\bibliography{ref}
\bibliographystyle{acl_natbib}

\end{document}